\documentclass{article}

\usepackage{arxiv}

\usepackage[utf8]{inputenc} % allow utf-8 input
\usepackage[T1]{fontenc}    % use 8-bit T1 fonts
\usepackage{hyperref}       % hyperlinks
\usepackage{url}            % simple URL typesetting
\usepackage{booktabs}       % professional-quality tables
\usepackage{amsfonts}       % blackboard math symbols
\usepackage{nicefrac}       % compact symbols for 1/2, etc.
\usepackage{microtype}      % microtypography
\usepackage{lipsum}
\usepackage{graphicx}
\graphicspath{ {./images/} }

\title{STREAMLINE: A Simple, Transparent, End-To-End Automated Machine Learning Pipeline Facilitating Data Analysis and Algorithm Comparison}

\author{
Ryan J. Urbanowicz \\
  Department of Computational Biomedicine\\
  Cedars Sinai Medical Center\\
  Los Angeles, CA 90048 \\
  \texttt{ryan.urbanowicz@cshs.org} \\
   \And
 Robert Zhang \\
  University of Pennsylvania\\
  Philadelphia, PA 19104\\
  \texttt{robertzh@wharton.upenn.edu} \\
  \And
 Yuhan Cui \\
  University of Pennsylvania\\
  Philadelphia, PA 19104\\
  \texttt{yuhan.cui@pennmedicine.upenn.edu} \\
    \And
 Pranshu Suri \\
  University of Pennsylvania\\
  Philadelphia, PA 19104\\
  \texttt{yuhan.cui@pennmedicine.upenn.edu} \\
}

\begin{document}
\maketitle
\begin{abstract}
Machine learning (ML) offers powerful methods for detecting and modeling associations often in data with large feature spaces and complex associations. Many useful tools/packages (e.g. scikit-learn) have been developed to make the various elements of data handling, processing, modeling, and interpretation accessible. However, it is not trivial for most investigators to assemble these elements into a rigorous, replicatable, unbiased, and effective data analysis pipeline. Automated machine learning (AutoML) seeks to address these issues by simplifying the process of ML analysis for all. Here, we introduce STREAMLINE, a simple, transparent, end-to-end AutoML pipeline designed as a framework to easily conduct rigorous ML modeling and analysis (limited initially to binary classification). STREAMLINE is specifically designed to compare performance between datasets, ML algorithms, and other AutoML tools. It is unique among other autoML tools by offering a fully transparent and consistent baseline of comparison using a carefully designed series of pipeline elements including: (1) exploratory analysis, (2) basic data cleaning, (3) cross validation partitioning, (4) data scaling and imputation, (5) filter-based feature importance estimation, (6) collective feature selection, (7) ML modeling with `Optuna' hyperparameter optimization across 15 established algorithms (including less well-known Genetic Programming and rule-based ML), (8) evaluation across 16 classification metrics, (9) model feature importance estimation, (10) statistical significance comparisons, and (11) automatically exporting all results, plots, a PDF summary report, and models that can be easily applied to replication data.
\end{abstract}

% keywords can be removed

\section{Introduction}
\label{sec:1}
Machine learning (ML) has become a cornerstone of modern data science with applications in countless research domains including biomedical informatics, a field synonymous with noisy, complex, heterogeneous, and often large-scale data (i.e. `big-data') \cite{thornton2004genetics,luo2016big,rauschert2020machine}. Surging interest in ML stems from its potential to train models that can be applied to make predictions as well as discover complex multivariate associations within increasingly large feature spaces (e.g. `omics' data as well as integrated multi-omics data) \cite{diao2018biomedical}. Increased access to powerful computing resources has fueled the practicality of these endeavors \cite{elsebakhi2015large}. As a result, a wealth of ML tools, packages, and other resources have been developed to facilitate implementation of custom ML analyses. One popular and accessible example includes the scikit-learn library built with the Python programming language \cite{pedregosa2011scikit}. Packages such as scikit-learn focus on facilitating the use of individual elements of an ML analysis pipeline (e.g. cross validation, feature selection, ML modeling). However, `how' these elements are brought together is generally left up to the practitioner leading to significant variability in how ML analyses are conducted, even when dealing with similar data. Guidelines for conducting ML analysis largely exist as a community knowledge pool of individual `potential pitfalls' and `best practices' \cite{smialowski2010pitfalls,luo2016guidelines,garreta2017scikit,vieira2020step,uccar2020effect,riley2019three,heil2021reproducibility,greener2022guide}. Most ML research tends to focus on improving or adapting individual methods for a given data type, task, or domain of application. Surprisingly few works have focused on how to effectively and appropriately assemble an ML pipeline in its entirety or provide accessible, easy to use examples of how to do so. For those coming from outside domains of expertise it can be daunting to know where to start.

In recent years, automated machine learning (i.e. AutoML) has emerged as a field of research, producing a number of strategies and tools aiming to facilitate and optimize the process of conducting machine learning data analyses \cite{hutter2019automated}. Specific AutoML tools differ based on (1) which elements of an ML analysis pipeline they automate, e.g. feature selection, model selection, and/or hyperparameter optimization, (2) whether they seek to automate the design of the pipeline itself, i.e. what elements to include and in what order (e.g. TPOT which conducts pipeline optimization using genetic programming \cite{olson2019tpot}), and (3) the types of data or application domains to which they were designed to be applied, e.g. tabular data vs. images, discrete vs. quantitative outcomes, etc. \cite{truong2019towards,chauhan2020automated,waring2020automated}. What makes most AutoML tools similar is that they focus on returning the single or subset of best results to the user. From the perspective of ML modeling this constitutes returning the model yielding the best performance based on a target evaluation metric. The advantage of AutoML tools to date includes greater accessibility, reduced tedium and researcher time in developing ML analysis pipelines, and offering potentially better optimization of modeling in contrast with manually conducted analyses. However, AutoML tools can be extremely computationally expensive and they still cannot guarantee optimal performance, or even necessarily better performance than a pipeline manually developed by an expert \cite{truong2019towards,fabris2019analysing}. Furthermore, AutoML tools are each inherently limited by the specific algorithms they implement and automate, and they generally do little to educate or actively engage users with respect to the process. This could potentially create new opportunities for ML misuse when practitioners are less engaged in the design analyses, particularly with respect to potential sources of bias or assumptions introduced by either the study design/data source, or the specific ML analyses conducted. 

As such, there are many opportunities for future AutoML to answer questions such as: (1) Can automated ML pipeline assembly reliably give comparable or better performance than analyses conducted by experts? (2) What are the specific strengths and weaknesses of different ML algorithms in tasks with different data types, data sizes, and underlying problem complexity? (3) Can we engineer AutoML tools to be smarter? Either learning from past analysis experience of the tool to conduct analyses more efficiently and effectively in subsequent runs (e.g. PennAI \cite{la2021evaluating}), or better constrain open ended AutoML search using human expertise on best practices and the use of specific algorithms, (4) Is there some subset of ML modeling algorithms that could be reliably applied more efficiently that represents a well-balanced cross section of algorithm strengths and weaknesses rather than applying all available algorithms? (5) How do we optimize ML interpretability? (6) How do we address covariates in ML analyses? (7) What is the best way to construct an ML analysis pipeline in different contexts? and (8) What aspects of a complete ML analysis pipeline can be reliably automated vs. which are best conducted using domain expertise?

With these questions in mind, we developed a simple, transparent, end-to-end, automated machine learning pipeline (STREAMLINE) which focuses exclusively on binary classification in tabular data in this initial implementation. Unlike most other AutoML tools, STREAMLINE is designed as a framework to rigorously apply and compare a variety of ML modeling algorithms in a carefully designed and standardized manner. STREAMLINE adopts a fixed series of purposefully selected ML analysis pipeline elements in line with data science best practices. It seeks to automate any domain-generalizable elements of an ML analysis pipeline including: exploratory analysis, basic data cleaning, cross-validation (CV) partitioning, data scaling, missing-value imputation, pre-modeling feature importance estimation, feature selection, ML modeling (including hyperparameter optimization within 15 scikit-learn \cite{garreta2017scikit} compatible ML algorithms that can be applied to binary classification), evaluation across 16 classification metrics, model feature importance estimation, generating and organizing all models, exporting publication-ready plots, and results, statistical significance comparisons across ML algorithms and analyzed datasets, generation of a summary report of settings and key results, and easy application and evaluation of all trained models to replication data. Of the 15 modeling algorithms currently included in STREAMLINE, 11 are popular, well established ML algorithms, 3 are rule-based evolutionary machine learning (RBML) algorithm implementations currently being developed by our group (i.e. eLCS, XCS, and ExSTraCS), and the last is a scikit-learn compatible implementation of genetic programming (GP) adapted for symbolic classification. We included the RBML algorithms to demonstrate how STREAMLINE can be used as a framework to test and compare new algorithms easily to other widely known standards. Further, the GP algorithm was included in part due to the relative ease of directly interpreting the models it produces in contrast with most other ML modeling algorithms, as well as to highlight GP as an accessible approach to modeling within the broader machine learning community. 

The design of STREAMLINE focused on (1) overall automation, (2) avoiding or allowing easier detection of bias, (3) optimizing modeling performance, (4) ensuring reproducibility, (5) capturing complex associations in data (e.g. feature interactions), (6) enhancing interpretability of output, (7) offering different use modes for varying levels of user computing experience, (8) facilitating pipeline output re-use to conduct further analyses and (9) making it easier to extend STREAMLINE in the future to include other algorithms or automated pipeline elements. Overall, the purpose of STREAMLINE is not to claim a `best way' to conduct an ML or AutoML analysis, but rather to serve as a baseline framework with which to easily conduct a straightforward but rigorous ML modeling analysis over one or more target datasets, as well as to facilitate interrogation of the questions presented earlier.

In the following sections we detail the automated elements of the STREAMLINE AutoML tool and provide an overview of (1) how it can be applied, (2) assumptions for use, and (3) output. Next we include a demonstration of STREAMLINE applied to a well-known benchmark dataset from the UCI repository \cite{Dua:2019} as well as a variety of simulated genomics datasets with distinct underlying patterns of associations, and multiplexer ML benchmark datasets. Lastly, we discuss future directions for STREAMLINE as a tool to support data analysis, ML algorithm and AutoML tool development, and encourage a better understanding of pipeline assembly for conducting effective and reliable ML analysis. 

\section{Methods}
\label{sec:2}
In this section we begin by detailing the automated elements of the STREAMLINE AutoML tool, discussing how it can be used, the assumptions it makes, and the output it produces. Then we will describe the datasets employed to test and demonstrate the efficacy of STREAMLINE. STREAMLINE is currently available at the following GitHub repository:  \url{https://github.com/UrbsLab/STREAMLINE} \cite{streamline}. We refer readers to the README of this repository for detailed instructions regarding (1) dataset formatting requirements, (2) installation instructions, (3) pipeline run parameters, and (4) different use modes. 

\subsection{STREAMLINE}
\label{subsec:2.1}
\begin{figure}[t!]
    \centering
    \includegraphics[width = \textwidth]{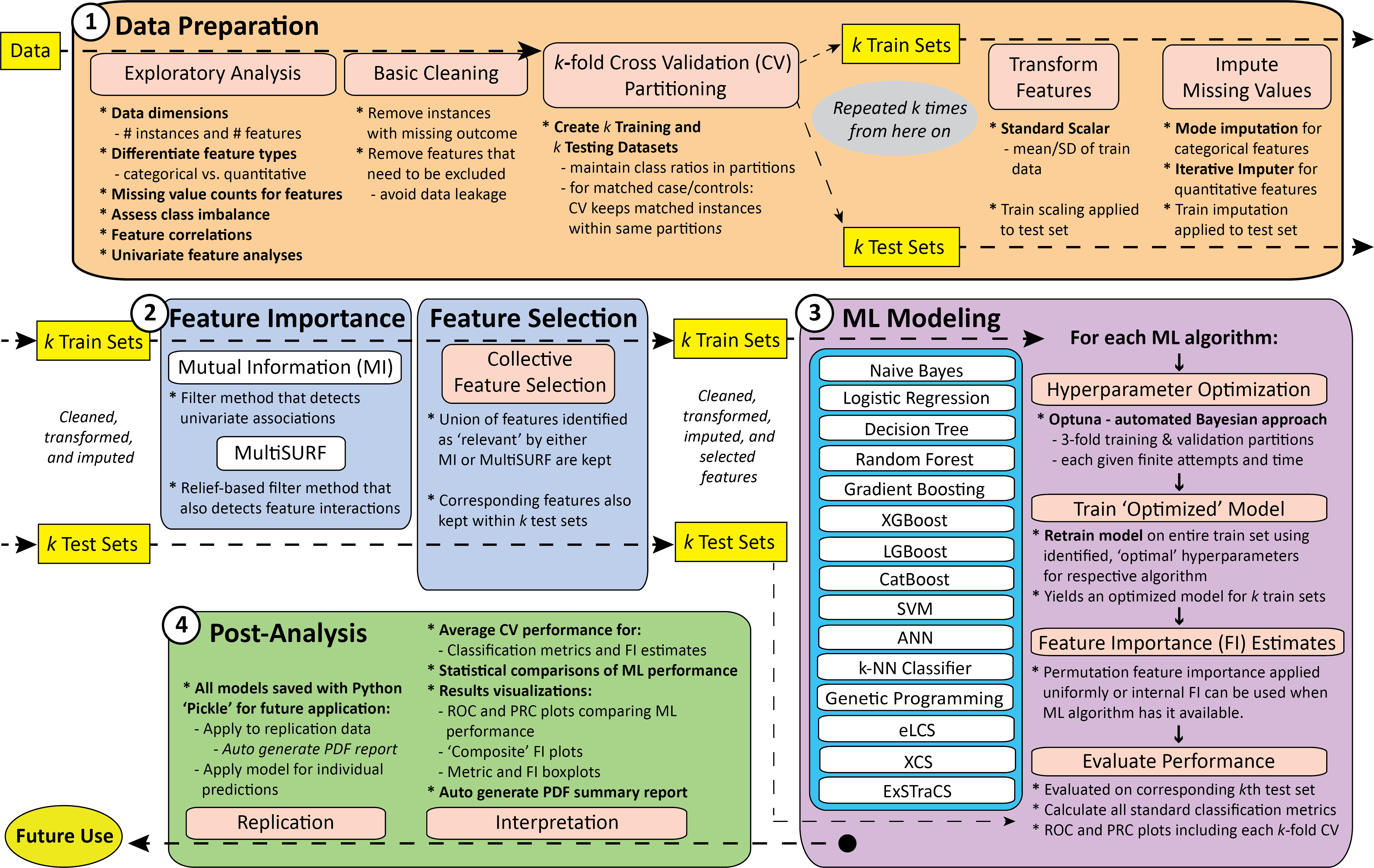}
    \caption{STREAMLINE Schematic. For each target dataset, STREAMLINE proceeds through a fixed series of pipeline elements generalized into the 4 stages above. Arrows indicate pipeline flow through each, with arrows leaving stage 1 corresponding to those entering stage 2. Target data and subsequently generated training and testing datasets are indicated by yellow boxes. Specific elements of the pipeline are indicated by pink boxes and specific algorithms are indicated by white boxes. Key details of pipeline functionality are noted.}
    \label{fig:my_label1}
\end{figure}

Figure \ref{fig:my_label1} provides a schematic overview of the elements included in the STREAMLINE tool. This schematic identifies key individual elements of the pipeline organized into 4 general stages: (1) data preparation, (2) feature importance estimation and selection, (3) ML modeling, and (4) post-analysis. The overall arrangement of pipeline elements was designed to avoid bias and data leakage, as well as to maximize parallelizability when run on a computing cluster. 

Below we detail the specific elements and reasoning behind their incorporation within each stage. Throughout this description we provide select figures illustrating STREAMLINE output when applied to an example simulated genomics dataset with 100 single nucleotide polymorphism (SNP) features and 1600 instances including 2 heterogeneously independent 2-way predictive interactions described in section \ref{subsec:2.2}. In practice, STREAMLINE offers recommended default run parameters that should work well for many analyses, however we encourage users to explore available pipeline run parameters to see how it could be more flexibly adapted to different studies \cite{streamline}. 

%Figure 1

\subsubsection{Data Preparation}
\label{subsubsec:2.1.1}
This STREAMLINE stage focuses on helping users understand their data as well prepare the data for downstream modeling. 
\\ \\
\textbf{Data Input:} STREAMLINE takes a folder including one or more properly formatted `target' datasets as input for a single `experiment', i.e. a complete run of the pipeline. This allows users to compare modeling performance across multiple datasets that might constitute scenarios such as: (1) the original dataset with or without covariates, to evaluate their impact, (2) datasets that include different subsets of available features, to evaluate the impact of explicitly including or excluding those features, or (3) datasets constituting different underlying challenges for ML modeling. Key requirements for dataset formatting include: numerical encoding of data values, a header of column labels, a unique and consistent identifier for missing data values, and having labeled data with a binary class outcome. All available instances can be included in these input data, or if a sufficient number of data instances are available, a random portion should ideally be withheld for replication analysis using STREAMLINE's `apply-model' phase in post-analysis. 
\\ \\
\textbf{Exploratory Analysis:} STREAMLINE summarizes each `target' dataset using summary statistics and plots including: (1) feature, instance, and missing value counts, (2) class balance, (3) feature correlation, and (4) univariate analyses. STREAMLINE automatically attempts to differentiate features that should be treated either as categorical or quantitative based on a (user defined) cutoff of unique value counts. Users can also manually specify which features to be treated as categorical. Depending on this distinction, appropriate univariate statistical tests and plots will be generated for each feature examining its relationship with outcome across the entire `target' dataset. Notably, in this first release of STREAMLINE the distinction of categorical vs. quantitative features is only utilized in the first two stages. Scikit-learn modeling will treat all features as if they were quantitative by default. Users that wish to ensure that categorical features are treated as such in modeling should employ one-hot-encoding prior to running STREAMLINE. This element will be automated in a future release, however we currently leave this decision up to users.
\\ \\
\textbf{Basic Cleaning:} Data cleaning is a notoriously difficult element to automate, and often requires many analysis-specific decisions. STREAMLINE data cleaning limits itself to removing instances that are missing an outcome value, as well as removing any features specified by the user that should be excluded (e.g. to prevent data leakage). STREAMLINE does not automatically remove outliers, as this is a decision best left to the user based on domain knowledge.  
\\ \\
\textbf{k-fold Cross Validation:} STREAMLINE employs k-fold cross validation (CV) prior to completing feature transformation, imputation, feature selection, or modeling to avoid any data leakage where information from the testing set is gleaned prior to final model evaluation. Users can choose from 3 CV strategies: (1) stratified, to ensure class balance within each partition, (2) random, and (3) matched, which allows users to keep specific groups of instances together within partitions \cite{krstajic2014cross}.  Matched CV partitioning allows users to properly analyze covariate-matched datasets to help control for covariate effects in ML analyses \cite{linden2016using}. STREAMLINE uses 10-fold stratified CV by default. From here and through modeling, all described elements of STREAMLINE are conduced within each of the k-fold partitions. 
\\ \\
\textbf{Impute Missing Values:} Missing data values presents practitioners with difficult decisions. Removing instances or features with missing values has the obvious drawbacks of reduced sample size and possibly removing relevant features, respectively. Some algorithm implementations are designed to handle missing values as neutrally as possible, which is likely the safest option, i.e. least likely to introduce unwanted bias \cite{white2010avoiding}. However STREAMLINE utilizes a number of scikit-learn ML modeling algorithm implementations that cannot handle missing values. As such, we view imputation here as a `necessary evil' that will afford us the opportunity to more easily compare a broader range of widely used ML modeling algorithms. STREAMLINE currently employs simple mode imputation for categorical variables to ensure the same set of possible categorical values are used to replace missing ones, and employs either iterative imputation \cite{buuren2011mice} or mean imputation as options for quantitative features. Iterative imputation considers the context of all features in the dataset, while both mode and mean imputation simply use information from the target feature. While iterative imputation is likely to perform better, it can be computationally expensive to run and take up a great deal of disk space to save the imputer object for future model applications when applied to large datasets. In STREAMLINE, imputation is first conducted in the training data partition, and then the same imputation decisions learned in training are applied directly to the testing partitions, or later to any replication data when using `apply-model'. Imputation is conducted prior to data scaling as the imputation could influence the correct center and scale to use. 
\\ \\
\textbf{Transform Features:} Feature transformation can play an important role in the ability of specific ML modeling algorithms to effectively detect and model different patterns of association \cite{kusiak2001feature}. Generally speaking, more sophisticated ML algorithms can still effectively model associations without employing feature transformation. One exception to this is feature scaling, where feature values across the dataset are scaled to have the properties of a standard normal distribution with a mean of zero and a standard deviation of one. While not critical for all algorithms (e.g. decision trees), algorithms such as support vector machines (SVM), k-nearest neighbors (KNN), logistic regression (LR), and artificial neural networks (ANN) require the data to be scaled to ensure effective training and/or proper interpretation of the resulting models, i.e. feature importance estimation. STREAMLINE employs scikit-learn's `standard scalar' to all datasets, performed first on the training data partition, and then applying the same learned scalar to testing or downstream replication data. 

\subsubsection{Feature Importance and Selection} 
This STREAMLINE stage focuses on providing a pre-modeling estimation of feature importance as well as seeking to conservatively remove features with no indication of being informative. We organized this stage as being separate from the rest of data processing, to reflect how this is a phase that can be viewed as part of both processing and modeling. It is an opportunity to gain further data insight about patterns in the data prior to modeling, as well as a potentially critical bottleneck in analyses dealing with large feature spaces, with many possible ways to go about it.

\begin{figure}[t!]
    \centering
    \includegraphics[width = \textwidth]{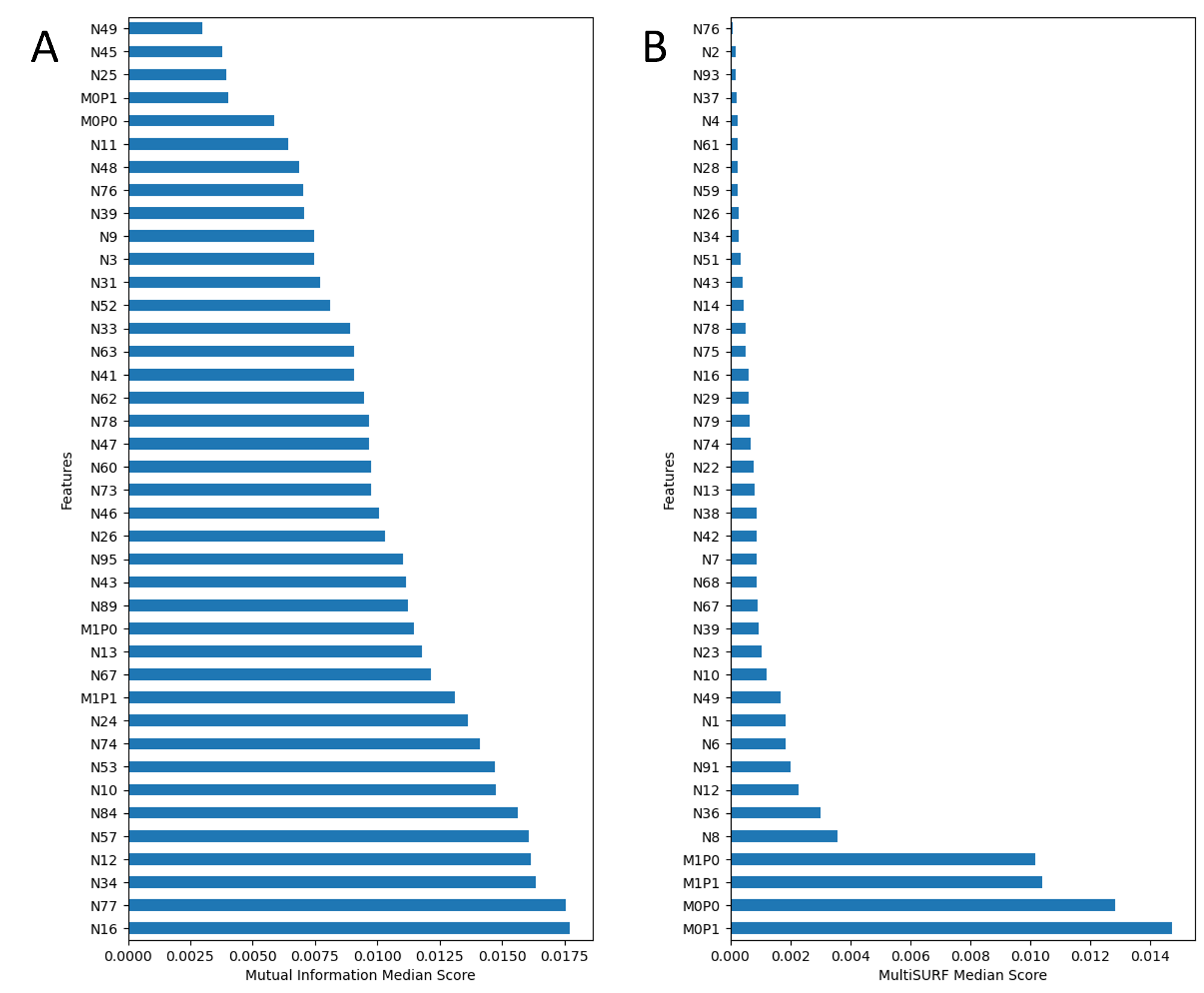}
    \caption{Median feature importance results across 10-fold CV on simulated SNP data. The left barplot gives MI scores and the right barplot gives MultiSURF scores. Simulated predictive feature names start with an `M' while non-predictive features start with an `N'. Note how MultiSURF perfectly prioritizes the four features simulated with epistatic and heterogenous patterns of association while MI does not.}
    \label{sim_fi}
\end{figure}

Notably, this first release of STREAMLINE does nothing to remove perfectly or highly correlated features from the data. We suggest users remove all but one feature of any fully correlated feature sets, and consider further removal of highly correlated features based on the needs of the problem at hand. 
\\ \\
\textbf{Feature Importance Estimation:} Prior to modeling it can be valuable to examine the estimated feature importance of features in the dataset. This can facilitate downstream interpretation, by providing something to compare model feature importance estimates to. Feature importance estimation can be based on a variety of feature selection (FS) strategies. While FS can take place in parallel with ML modeling (i.e. wrapper-based or embedded FS \cite{urbanowicz2018relief}), this can be computationally expensive, difficult to automate, and can identify feature subsets that are biased by the modeling algorithms being applied. Therefore this pipeline focuses on utilizing filter-based FS methods, which are typically fast and can be combined with any downstream ML method. Unfortunately, most filter-based FS methods are insensitive to complex feature interactions. Ultimately we are concerned with detecting both simple and potentially complex associations. As such, STREAMLINE implements two FS algorithms, i.e. mutual information (MI) \cite{peng_feature_2005} and MultiSURF \cite{urbanowicz2018benchmarking} (a Relief-based FS algorithm).  MI is proficient at evaluating univariate associations between a feature and outcome, while MultiSURF has been demonstrated to be sensitive to not only univariate associations but also both 2 and 3 way feature interactions (even in the absence of univariate associations).
\\ \\
\textbf{Feature Selection:} A key factor in the scalability of any AutoML tool will be the number of instances and features in the target data. Feature selection is an element of an ML analysis pipeline that, in some cases, can be conducted directly by ML modeling algorithms. However, in general we can expect ML modeling algorithms to perform best if irrelevant or redundant features have been removed first. A critical mistake in conducting feature selection prior to ML modeling is to accidentally remove relevant features using over-simplistic selection methods such as statistical tests or algorithms that only account for simple univariate or linear associations. Previous research has suggested that FS conducted by an ensemble of methodologies provides an effective approach to avoid accidentally removing relevant features. As such, STREAMLINE adopts `collective feature selection' that calls for the use of more than one FS methodology in making the determination as to whether a given feature should be retained or removed \cite{verma2018collective}. If either MI or MultiSURF finds evidence that a feature may be informative, that feature is retained, otherwise it is removed. Both algorithms estimate feature importance scores for every feature where a score larger than $0$ is considered to be potentially informative (and the feature is retained). In practice, irrelevant features can have values $> 0$ due to random chance in feature values, but we are conservative in selecting features to avoid removing features with potentially small but meaningful effects. Users can set STREAMLINE to keep all features regardless of feature importance estimates, or set a cap on the maximum number of features to allow in each dataset prior to modeling. If the latter is selected, STREAMLINE will first drop any features with a score $<=$ 0 then alternate between each FS algorithm taking the unique, top-scoring features until the maximum is reached. Figure \ref{sim_fi} presents median feature importance scores (over all CVs) for the two FS algorithms applied to the aforementioned example data. This highlights the potential value of collective feature selection, given that if we had only used MI, some of the 4 predictive features (M0P0, M0P1, M1P0, M1P1) may have been removed based on how the maximum features to keep was set prior to modeling. 

%Figure 2

MultiSURF scales quadratically with the number of training instances, therefore this notebook includes a user parameter to control the maximum number of randomly selected instances used by MultiSURF during scoring (set to 2000 by default to avoid very large run-times).

Of note, it has been previously demonstrated that when there are 10,000 features or more, MultSURF begins to fail to properly rank features involved in a pure interaction. However, Relief wrapper methods such as TuRF \cite{moore2007tuning} have been proposed that can boost ranking success by iteratively filtering out the poorest scoring features over subsequent runs of the Relief-based algorithm. STREAMLINE includes the option to apply TuRF, which we recommend using in datasets with $>$ 10,000 features. 

FS takes place independently within each of the \emph{k} training sets meaning that it is possible for each set to have a different subset of features available for model training. Notably, STREAMLINE exports the training and testing sets generated after CV partitioning, imputation, scaling, and feature selection so users can easily apply other algorithms and analyses to these same sets outside the pipeline. 

\subsubsection{Modeling} 
The heart of STREAMLINE is ML modeling itself. ML includes a large family of algorithmic methodologies that differ with respect to knowledge representation and inductive learning approaches as well as to the types of data and associations they are most effectively applied to and how interpretable their respective models will be. This first release of STREAMLINE includes a total of 15 scikit-learn compatible implementations of ML classification algorithms that can be individually utilized or left out. This set was selected based on their general popularity, ease of pipeline integration, and to offer variety with respect to model representation and learning approaches. These include: naive bayes (NB) \cite{scikit-nb}, logistic regression (LR) \cite{scikit-lr}, decision tree (DT) \cite{scikit-dt}, random forest (RF) \cite{scikit-rf}, gradient boosting trees (GB) \cite{scikit-gb}, extreme gradient boosting (XGB) \cite{xgboost}, light gradient boosting (LGB) \cite{lgboost}, catboost (CGB) \cite{catboost}, support vector machines (SVM) \cite{scikit-svm}, artificial neural networks (ANN) \cite{scikit-ann}, k-nearest neighbors (KNN) \cite{scikit-knn}, genetic programming (GP) symbolic classifier using `gp-learn' \cite{gp-learn}, educational learning classifier system (eLCS) \cite{zhang2020scikit,scikit-elcs}, `X' classifier system (XCS) \cite{scikit-xcs}, and an extended supervised tracking and classifying system (ExSTraCS) \cite{urbanowicz2015exstracs,zhang2021lcs,scikit-exstracs}. ExSTraCS is a RBML algorithm developed by our research group which combines evolutionary learning with a `piece-wise' knowledge representation comprised of a set of IF:THEN rules. Unlike most ML methods, many RBML algorithms learn iteratively from instances in the training data and adapt a population of IF:THEN rules to cover unique parts of the problem space making them particularly well suited to capturing heterogeneous patterns of association \cite{urbanowicz2015exstracs}. RF, GB, XGB, LGB, and CGB are all decision tree based algorithms and are currently among the most popular and successful ML algorithms within the ML research community. The implementation of GP we selected for inclusion is scikit-learn compatible, accessible, simple to use. We include GP in STREAMLINE to highlight it as one of the few directly interpretable modeling algorithms, however, as a simple implementation, we do not expect its performance to reflect what other more sophisticated and modern GP algorithms are capable of. 

Notably, many algorithms have other implementation options available which might be better tuned to specific problem domains, or yield better performance in general. As such any ML comparison results from STREAMLINE, or any other AutoML tool that relies on specific implementations, should not be casually interpreted as a demonstration of the general superiority of one algorithm over another. By default, STREAMLINE will run all algorithms with the exception of eLCS and XCS which are still actively being developed and have known issues identified in early testing of this pipeline. This also highlights how STREAMLINE can be an effective development tool to test, debug, and evaluate new ML algorithm implementations. 

STREAMLINE intentionally does not automate 'algorithm selection' as it seeks to present and compare ML algorithm modeling performance in as transparent and unbiased a manner as possible across a wide variety of evaluation metrics. This is intended to give users greater insight with respect to findings, and greater flexibility with respect to how AutoML results can be utilized. 
\\ \\
\textbf{Hyperparameter Optimization:} According to current best practices, the first step in ML modeling is to conduct a hyperparameter optimization sweep, a.k.a. `tuning' \cite{schratz2019hyperparameter}. Hyperparameters, refer to the run parameters of a given ML algorithm that controls its functioning. Too often, ML algorithms are applied using their 'default' hyperparameter settings. This can lead to unfair ML algorithm comparisons, and a missed opportunity to obtain the best performing model possible. Optimization effectively 'tries out' different hyperparameter settings on a subset of the training instances, ultimately selecting those yielding the best performance with which to train the final model. The first consideration, is what hyperparameters to explore for each algorithm, as well as the range or selection of hyperparameter values to consider for each. As there is no clear consensus, we have surveyed a number of online sources, publications, and consulted colleagues in order to select the wide variety of hyperparameters and value ranges incorporated into STREAMLINE that are suited to binary classification. These hyperparameters and value options for each algorithm are detailed in the STREAMLINE Jupyter Notebook, as well as hard coded in the associated ModelJob.py script \cite{streamline}. For example, STREAMLINE considers 8 hyperparameters for the optimization of the RF algorithm including `number of estimators', i.e. the number of trees in the `forest', with potential values ranging from 10 to 1000. Users applying the Jupyter Notebook run mode can easily adjust the values considered in the hyperparameter optimization of each ML algorithm if desired, however we have set these options to be as broadly applicable as possible. 

The second consideration, is what approach to take in conducting the hyperparameter sweep. Common approaches include a `grid search' or a `random search' \cite{uppu2017tuning} which either exhaustively or randomly consider hyperparameter value combinations. STREAMLINE adopts an `automated' package called `Optuna', which applies Bayesian optimization to try and `intelligently' explore the specified hyperparameter space \cite{akiba2019optuna}. By default, a further 3-fold nested CV is conducted on the training dataset using a primary evaluation metric, where the user specifies a target number of optimization trials to complete and a maximum timeout to more equally limit the amount of run time each algorithm can spend on optimization. However in order for STREAMLINE to ensure complete reproducibility, particularly when run in parallel, the user must specify `None', for the Optuna timeout parameter, since variations in computing speed can lead to a different number of trials having been completed for each run.  By default, all algorithms utilize hyperparameter optimization with the exception of Naive Bayes (which has no hyperparameters) as well as eLCS, XCS, and ExSTraCS which are more computationally expensive, in general, and have relatively reliable default settings and reasonable guidelines to choose non-default hyperparameter settings as needed.  This likely puts ExSTraCS at somewhat of a disadvantage compared to other ML algorithms in this pipeline.

\begin{figure}[t!]
    \centering
    \includegraphics[width = \textwidth]{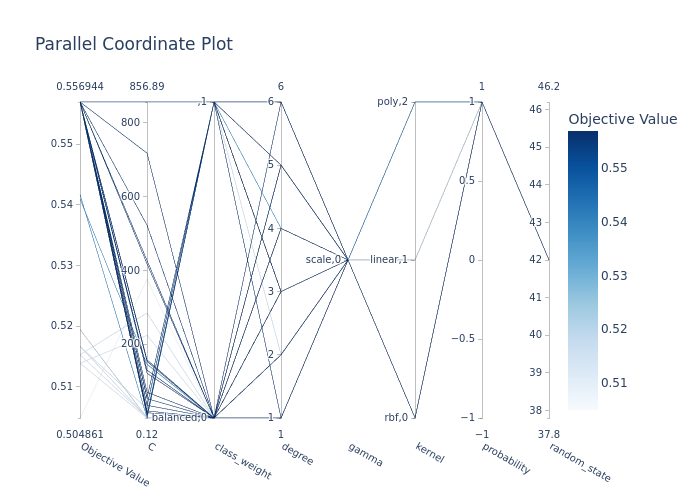}
    \caption{Optuna-generated visualization of the SVM algorithm hyperparameter sweep conducted on one CV partition of the example simulated genomics data.}
    \label{svm_optuna}
\end{figure}

STREAMLINE also provides users with the option to output hyperparameter sweep visualizations to illustrate how different settings combinations impact performance (see Figure \ref{svm_optuna}). Note how the sweep suggests that the 'kernel' hyperparameter (on the x-axis) performs poorly when set to `linear', performs better when set to `poly' and best when set to `rbf', i.e. radial basis function, when applied to this data.

%Figure 3

\textbf{Train `Optimized' Model:} Once `optimal' hyperarameters have been selected for each of the \emph{k} training datasets and ML algorithms, a `final' model is trained for each using the full training set and respective selected hyperparameters. The resulting models are stored as `pickled' Python objects so that they can be easily utilized later. 
\\ \\
\textbf{Model Feature Importance Estimation:} Next, this pipeline further examines feature importance, now from the perspective of each ML model. These estimates offer useful insights for high level interpretation of models. Specifically, which features were most important for making accurate predictions. Some algorithm implementations, including LR, DT, RF, GB, XGB, LGB, CGB, eLCS, XCS, and ExSTraCS have built-in strategies to report feature importance estimates after model training. In order to obtain feature importance estimates for the other 5 algorithms this pipeline implements permutation feature importance in scikit-learn. This strategy randomly permutes one feature value at a time in the testing data to evaluate its impact on the target performance metric. STREAMLINE uniformly utilizes permutation feature importance for all algorithms by default in order to provide a consistent metric for comparison. However users can chose to use built-in algorithm strategies whenever available, and permutation feature importance for the other 5 algorithms as an alternative.

Any feature that was removed by feature selection prior to modeling is given a model feature importance estimate of zero by default in the respective CV partition. 
\\ \\
\textbf{Evaluate Performance:} The last element in ML modeling is to evaluate model performance. It is essential to select appropriate evaluation metrics to fit the characteristics and goals of the given analysis. STREAMLINE ensures that a comprehensive selection of binary classification metrics and visualizations are employed that offer a holistic perspective of model performance. Earlier elements of this pipeline call for a target evaluation metric, i.e. hyperparameter optimization and model feature importance estimation. STREAMLINE uses \emph{balanced accuracy} by default, since this metric equally emphasizes accurate predictions within both classes. In addition to balanced accuracy this notebook calculates and reports the following 15 evaluation metrics: true positive count, true negative count, false positive count, false negative count, standard accuracy, F1-Score, sensitivity/recall, specificity, precision, receiver operating characteristic (ROC) area under the curve  (AUC), precision-recall curve (PRC) AUC, PRC average precision score (APS), negative predictive value, positive likelihood ratio, and negative likelihood ratio. STREAMLINE saves model performance metrics, feature importance estimates, and can output prediction probabilities (post-hoc) on testing or replication data as .csv files and 'pickled' objects (to facilitate future use) for each algorithm and CV partition. 

\subsubsection{Post-Analysis} 
\begin{figure}[t!]
    \centering
    \includegraphics[width = \textwidth]{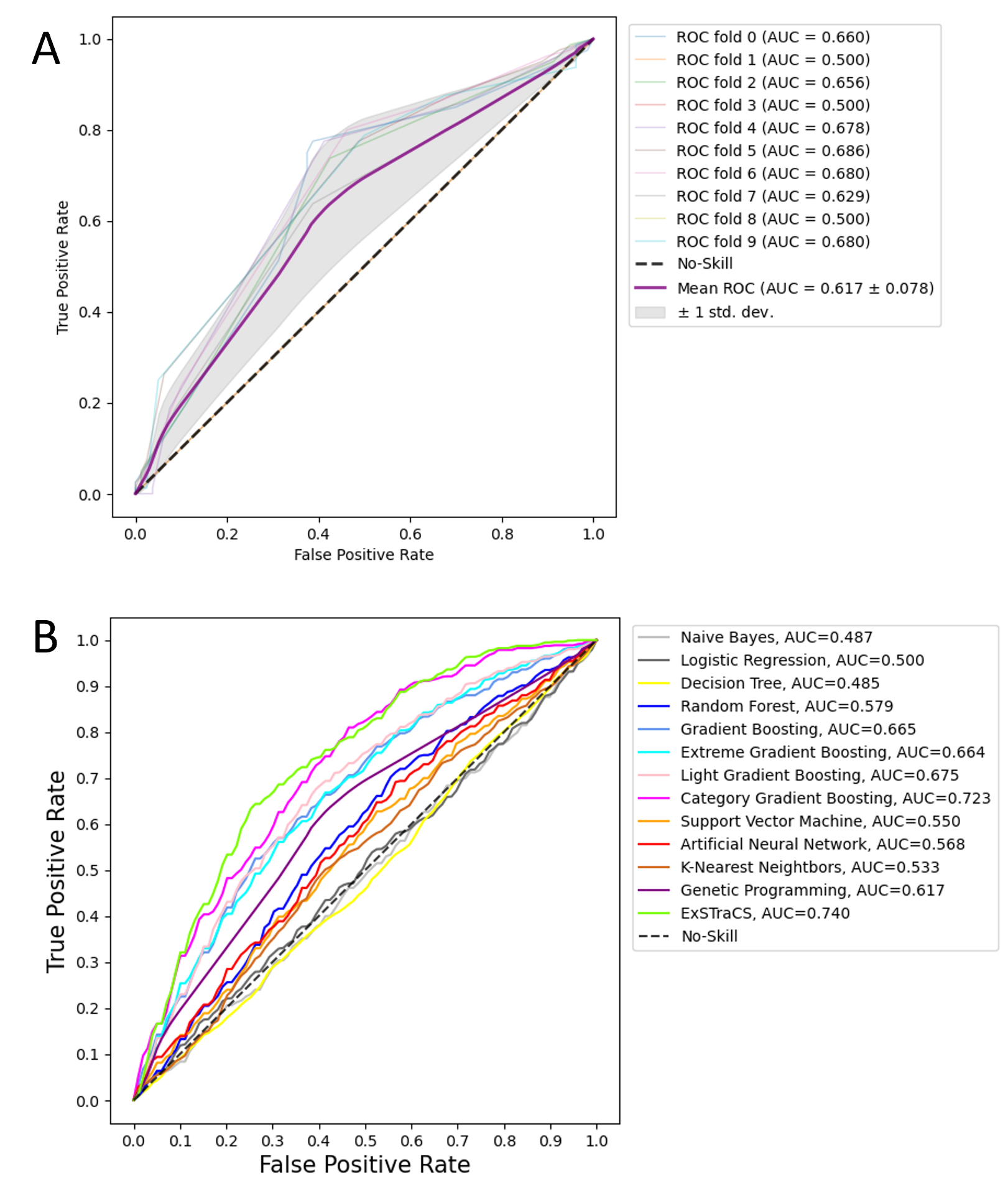}
    \caption{ROC plots for the example simulated genomics dataset: (A) for each of 10 CV models trained with the GP algorithm, and (B) for average performance of the 13 ML algorithms that STREAMLINE runs by default. Note that our rule-based ML algorithm, ExSTraCS performs at least as well as all other algorithms in this example despite not having the benefit of hyperparameter optimization. This illustrates how STREAMLINE may be applied to verify the efficacy of new ML algorithms.}
    \label{roc}
\end{figure}

\begin{figure}[t!]
    \centering
    \includegraphics[width = \textwidth]{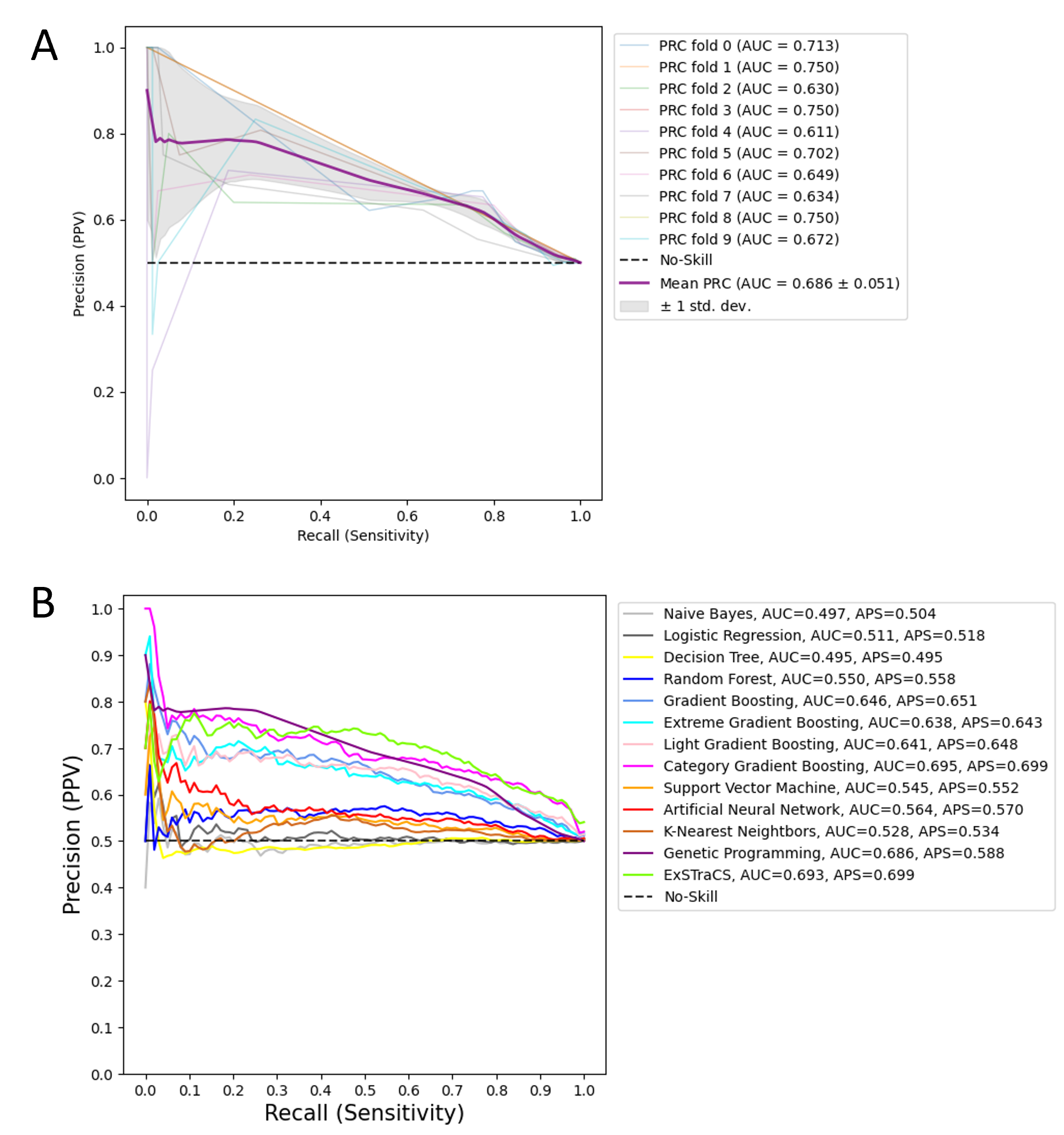}
    \caption{PRC plots for the example simulated genomics dataset: (A) for each of 10 CV models trained with the GP algorithm, and (B) for average performance of the 13 ML algorithms that STREAMLINE runs by default.}
    \label{prc}
\end{figure}

STREAMLINE post analysis calculates mean and median performance results (over all CV partition models), generates figures, conducts non-parametric statistical comparisons, generates a PDF summary report of key findings, and allows easy application of all trained models to available replication data or other future data to further evaluate model generalization capability. 
\\ \\
\textbf{Performance Summary:} STREAMLINE calculates CV means, medians, and standard deviations for all evaluation metrics and model feature importance scores.
\\ \\
\textbf{Figure Generation:} STREAMLINE automatically generates a wide variety of figures capturing performance and promoting model interpretability. For each algorithm, STREAMLINE pipeline outputs an ROC and PRC plot summarizing performance of each of the \emph{k} trained models along side the mean ROC or PRC, respectively. PRC plots are preferable to ROC plots when class imbalance is more extreme or if the practitioner is more concerned with making positive class predictions. This pipeline automatically sets the 'no-skill' lines of PRC plots based on the class ratios in the dataset. Additionally, ROC and PRC plots comparing average CV performance across all algorithms are generated. Figure \ref{roc} presents ROC plots and Figure \ref{prc} presents PRC plots illustrating comparison of (A) individual CV runs of the GP algorithm, and (B) averaged performance of all algorithms. 

%Figure 4

%Figure 5

\begin{figure}[t!]
    \centering
    \includegraphics[width = \textwidth]{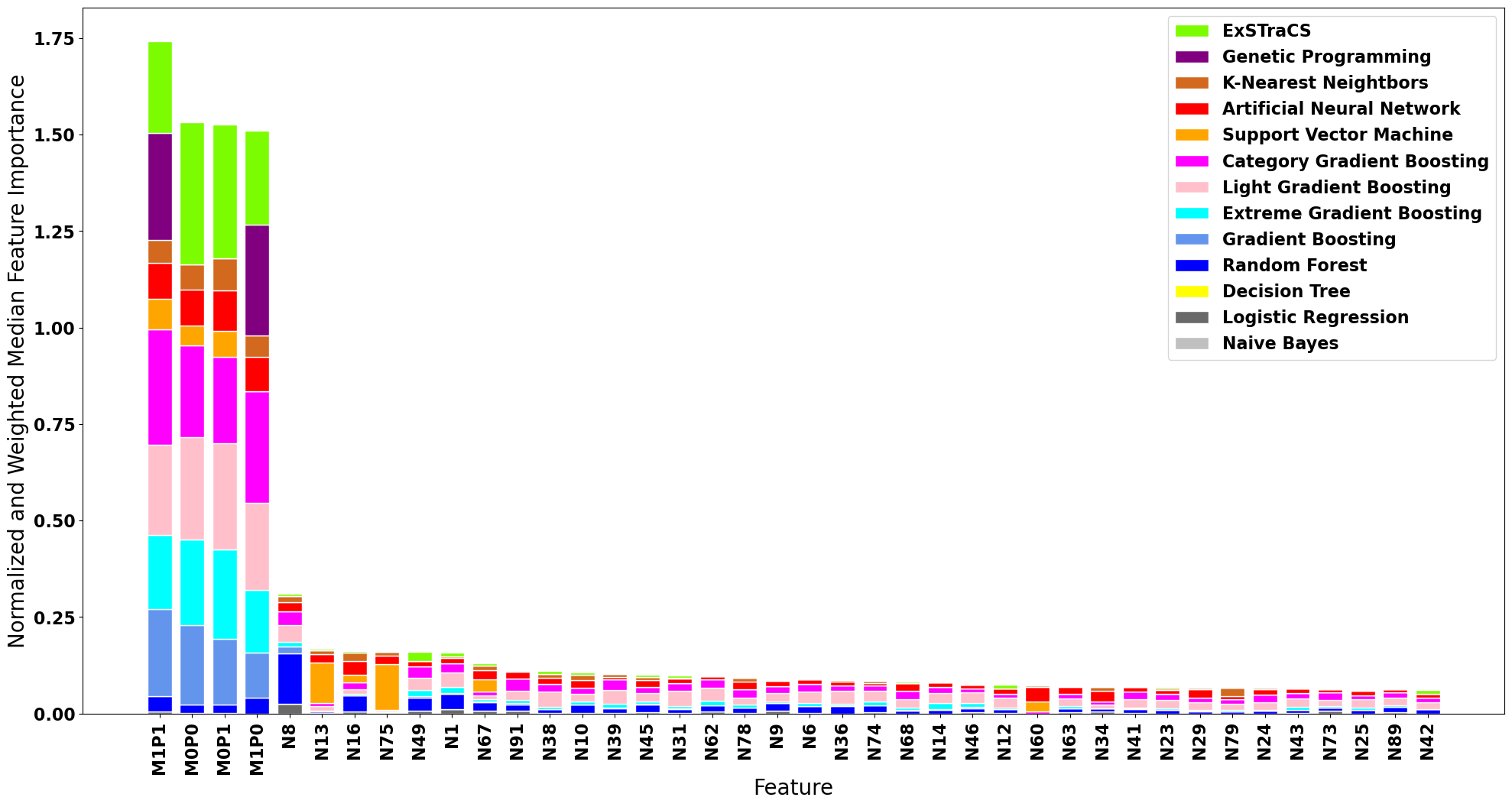}
    \caption{Composite feature importance bar plot (CFIBP) for the example simulated genomics dataset (top 40 features). Note how top performing algorithms correctly identify M0P0, M0P1, M1P0, and M1P1 as most important, but (in this data scenario) the RF algorithm incorrectly prioritizes N8, having performed less well in comparison.}
    \label{cfibp}
\end{figure}

STREAMLINE also automatically generates (1) boxplots for each evaluation metric comparing performance across all ML algorithms, (2) boxplots of top, ranked model feature importance estimates for each ML algorithm, and (3) histograms of feature importance score distribution for each ML algorithm (not shown here). STREAMLINE also generates proposed 'composite feature importance bar plots (CFIBP) that illustrate and summarize model feature importance consistency across all ML algorithms. The focal CFIBP generated normalizes feature importance scores within each algorithm between 0 and 1 and then weights the normalized scores by the median balanced accuracy or median ROC-AUC of the the respective algorithm so that those which do not perform as well have less impact on the visualization. Figure \ref{cfibp} gives the CFIBP generated for the example genomics dataset, illustrating the consensus of top performing algorithms in identifying the true predictive features. 

%Figure 6

If STREAMLINE is applied to multiple datasets in an `experiment' it will create boxplots for each metric comparing average results across all ML algorithms including lines indentifying how the performance of each ML algorithm changes. It will create similar boxplots for each ML algorithm and metric combination comparing CV runs from one dataset to the next. Examples of these dataset comparison figures are included in section \ref{sec:3}.

\begin{figure}[t!]
    \centering
    \includegraphics[width = 0.69\textwidth]{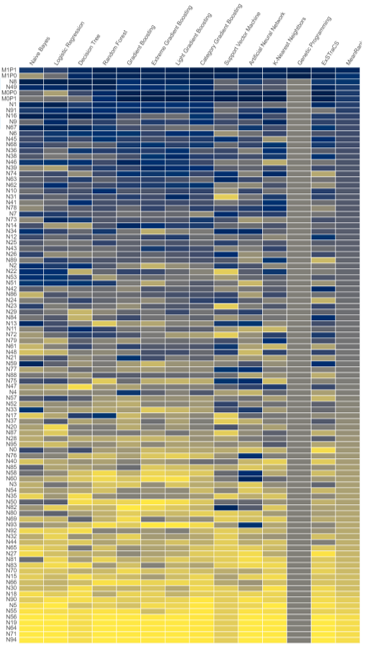}
    \caption{Model feature importance heat-map with average CV ranking across all ML algorithms. Dark blue denotes higher importance and bright yellow denotes low importance. This interactive heat-map can be opened as a web-page generated by one of the STREAMLINE `Useful Notebooks'. Placing the cursor over any cell (when opened as an `html') will reveal the algorithm, feature name, and rank of that feature for the given algorithm.}
    \label{firanking}
\end{figure}

\begin{figure}[h]
    \centering
    \includegraphics[width = \textwidth]{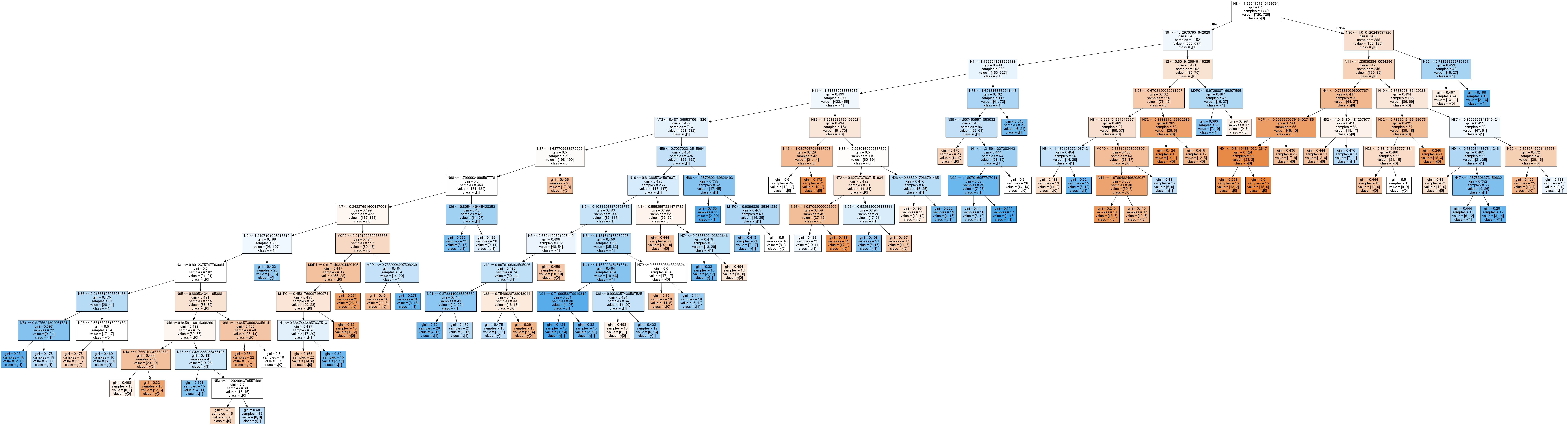}
    \caption{Direct visualization of the DT model with the highest CV testing ROC-AUC for the example simulated genomics dataset. As seen in Figure \ref{roc}, DT performs quite poorly on this dataset, and this overly complicated DT tree is struggling to detect any associations in the presence of heterogeneous feature interactions.}
    \label{dt_fig}
\end{figure}

\begin{figure}[h]
    \centering
    \includegraphics[width = 0.3\textwidth]{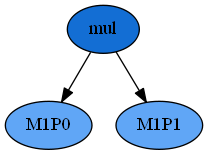}
    \caption{Direct visualization of the GP model with the highest CV testing ROC-AUC for the example simulated genomics dataset. As seen in Figure \ref{roc}, GP performs only moderately well on this dataset compared to some other algorithms. This simple and clearly interpretable GP tree is picking up on one of the two heterogeneous 2-way feature interactions simulated in the dataset suggesting that this GP implementation can tackle 2-way feature interactions, but struggles to model heterogeneous relationships.}
    \label{gp_fig}
\end{figure}

Included in the STREAMLINE GitHub repository \cite{streamline} is a folder of `Useful Notebooks', i.e. Jupyter Notebooks designed to do facilitate an array of additional tasks following the main pipeline analysis including (1) exporting prediction probabilities for training, testing, or replication data, (2) making custom modifications to key figures, (3) exploring and evaluating different decision thresholds beyond the standard 0.5 (4) generating a ranked interactive model feature importance heat-map (see Figure \ref{firanking}), and (5) generating direct visualizations of the most interpretable classification algorithms, i.e. DT and GP (see Figures \ref{dt_fig} and \ref{gp_fig}). 

%Figure 7

%Figure 8

%Figure 9

\textbf{Statistical Comparisons:} STREAMLINE is set up to automatically conduct non-parametric statistical comparisons (1) between ML algorithms using CV partitions, and (2) (if more than one dataset is being analyzed) between datasets for each individual ML algorithm as well among the best performing ML algorithms for every evaluation metric. In either scenario, STREAMLINE first applies Kruskal-Wallis one-way analysis of variance to determine if any ML algorithm or datasets yielded significantly different performance than the others for each evaluation metric. For any metric where a significant difference was observed, STREAMLINE follows up with pairwise Mann-Whitney U-tests and Wilcoxon Rank tests for that metric to identify which pairs of algorithms or datasets yielded significantly better or worse performance. 

\textbf{PDF Summary Report:} Given the wide range of output files generated and organized by STREAMLINE, it also automatically generates a formatted PDF summary report laying out all STREAMLINE run settings, a descriptive summary of key dataset characteristics, evaluation and feature importance plots, and metric results for each dataset, as well as Kruskal Wallis significance results comparing top algorithms for each metric/dataset combination, and runtime of all elements and specific ML algorithms. A report is generated for each STREAMLINE `experiment' as well as for any application of trained models to any additional replication datasets. 

\textbf{Application to Replication Data:} Once STREAMLINE has run an `experiment' all trained models can be evaluated on additional hold-out replication data to help verify model generalizability and globally pick a `best' model. This 'apply-model' phase of STREAMLINE automatically loads pickled models, as well as information on how to impute and scale the data in line with the original training data. This can be run on any number of additional replication datasets. If a user wishes to select a final model to put into practice as a predictive model in real-world applications such as clinical decision support, we recommend using the the results of applying all models to the replication data in order to choose it. This (1) ensures that all models are evaluated and compared using the same hold out data, (2) reduces opportunity for sample bias, and (3) as provides further rigor in the assessment of model generalizability. Differently, outside of STREAMLINE, users could create an ensemble model (as their final model) that combines the predictions of all CV models for a single algorithm, or that combines all models (algorithms and CV partitions) trained by the pipeline. Automating ensemble prediction as as STREAMLINE option will be included in a future release.

\subsubsection{Implementation and Run Modes}
STREAMLINE was implemented using Python 3 and a variety of well established Python packages including scikit-learn, pandas, scipy, numpy, optuna, and pickle. STREAMLINE has 4 available run modes designed to suit users with different levels of computing experience and analysis needs. We review these modes below, but direct users to the STREAMLINE README for specific instructions for installation and use \cite{streamline}.  
\\ \\
\textbf{Mode 1 (Google Colab Notebook):} This is the easiest way to run STREAMLINE, requiring no coding or computing environment experience, is using a Google Colab Notebook. This runs the pipeline serially using Google Colab and Google Cloud for computing. This is currently the slowest and most computationally limited option. It's ideal for those new to ML looking to learn or run simpler, smaller-scale analyses.
\\ \\
\textbf{Mode 2 (Jupyter Notebook):} For those with a little Python experience, and are familiar with installing Anaconda, and the commands to install other Python packages. This option offers users easy access to modify STREAMLINE run parameters and ML algorithm hyperparameter options. This option also runs serially, but gives users direct access to results and figures within the notebook as they are generated. 
\\ \\
\textbf{Mode 3 (Local Command Line):} For those with command line experience this run mode is likely preferable. The 4 stages of STREAMLINE, described above are set up to run (in sequence) as 11 possible command line phases optimized (in particular) for Mode 4 parallelization. This is detailed in \cite{streamline}.
\\ \\
\textbf{Mode 4 (Computing Cluster Parallelized Command Line):} For those with access to an LSF compatible Linux computing cluster, STREAMLINE is currently set up to efficiently parallelize its 4 stages as 11 compute phases. Programming savy users should be able to easily update the 'Main' Python scripts to parallelize STREAMLINE analyses using other distributed computing environments such as Amazon Web Services, Microsof Azure, Google Cloud or other local compute clusters. 

\subsection{Evaluation Datasets}
\label{subsec:2.2}
\subsubsection{HCC UCI Benchmark Data}
In this paper we first evaluated STREAMLINE performance first on the hepatocellular carcinoma (HCC) survival benchmark dataset from the UCI repository \cite{Dua:2019}. We chose the HCC dataset as the demonstration dataset, included with the STREAMLINE software, as it's small, i.e. 165 instances, and 49 features, thus quick to analyze, and because in exemplifies many of the standard data considerations/challenges including having (1) a mix of feature types (i.e. categorical and numerical), (2) about $10\%$ missing values, and (3) class imbalance with 63 patients who were deceased and 102 who survived. For the purposes of demonstration, we created as second dataset from this HCC data that removed two common covariates (i.e. age and sex) to show how STREAMLINE could be applied to multiple datasets at once in order to easily compare dataset performance in this context. We repeated this analysis twice in parallel using a fixed number of Optuna trials and no timelimit to confirm the replicability of STREAMLINE. To test the `apply-model' functionality of STREAMLINE we used the entire HCC dataset as a stand-in replication dataset, since a true replication dataset wasn't available and training instances were limited. 

\subsubsection{GAMETES Simulated Datasets}
Next, we evaluated 6 other genomics datasets simulated with GAMETES \cite{urbanowicz2012gametes}, each with a different underlying modeled pattern of association to illustrate how STREAMLINE can be utilized to evaluate the strengths and weaknesses of different ML algorithms in different contexts. Each datasets simulated 100 single nucleotide polymorphisms (SNPs) as features and included 1600 instances using a minor allele frequency of 0.2 for relevant features and the 'easiest' model architecture generated for each configuration using the GAMETES simulation approach \cite{urbanowicz2012predicting}. All irrelevant features were randomly simulated with a minor allele frequency between 0.05 and 0.5. We detail dataset differences here: (A) Univariate association between a single feature and outcome with 1 relevant and 99 irrelevant features and a heritability of 0.4, (B) Additive combination of 4 univariate associations with 4 relevant and 96 irrelevant features and heritability of 0.4 for each relevant feature, (C) Heterogeneous combination of 4 univariate associations, i.e. each univariate association is only predictive in a respective quarter of instances (with all else the same as dataset B), (D) Pure 2-way feature interaction with 2 relevant and 98 irrelevant features and heritability of 0.4, (E) Heterogeneous combination of 2 independant pure 2-way feature interactions with 4 relevant and 96 irrelevant features and heritability of 0.4 for each 2-way interaction (used in above sections for figure examples), (F) Pure 3-way feature interaction with 3 relevant and 97 irrelevant features and heritability of 0.2 (i.e. the most difficult dataset). 

When STREAMLINE conducts modeling in the datasets discussed so far, ExSTraCS, which was trained without a hyperparameter sweep, is set to run with hyperparameters (\emph{nu} = 1, rule population size =2K, and training iterations = 200K), where the \emph{nu} setting represents the emphasis on discovering and keeping rules with maximum accuracy. Previous work suggest that \emph{nu} = 1 is more effective in noisy data, while a \emph{nu} of 5 or 10 is more effective in clean data (i.e. no noise). 

\subsubsection{\emph{x}-bit MUX Benchmark Datasets}
Lastly, we applied STREAMLINE to 6 different \emph{x}-bit multiplexer (MUX) binary classification benchmark datasets (i.e. 6, 11, 20, 37, 70, and 135-bit) often utilized to evaluate ML algorithms such as RBML \cite{urbanowicz2015exstracs}. Like GAMETES dataset `E', these MUX datasets involve both feature interactions and heterogeneous associations with outcome, but differently involve binary features and a clean association with outcome. The `\emph{x}'-bit value denotes the number of relevant features underlying the association, and increasing values from 6 to 135 dramatically scale-up the complexity of the underlying pattern of association. Specifically, solving the 6-bit problem involves modeling 4 independent 3-way interactions, while the 135-bit problem involves modeling 128 independent 8-way interactions. In these datasets, all features are relevant to solving the problem, where a subset of features serve as `address bits` which point to one of the remaining `register bits'. The value of that corresponding register bit indicates the correct outcome (i.e. 0 or 1). 

In previous work, the ExSTraCS algorithm was the first ML demonstrated to directly model the 135-bit problem successfully \cite{urbanowicz2015exstracs}, in a dataset including 40K training instances, with hyperparameters (\emph{nu} = 10, rule population size =10K, and training iterations = 1.5 million). In this study, 6 to 135-bit datasets were generated with 500, 1000, 2000, 5000, 10000, and 20000 instances respectively, with 90$\%$ of each used by STREAMLINE for training (due to 10-fold CV). In previous work it was demonstrated that larger numbers of instances are required to solve or nearly solve increasingly complex MUX problems. In contrast with modeling on the noisy datasets, for the MUX datasets, ExSTraCS was assigned the following hyperparameter settings: \emph{nu} = 10, rule population size =5K, and training iterations = 500K). 

All analyses were conducted in parallel on an LSF compatible, Linux computing cluster with Anaconda3-2022.05-Linux-x86$\_$64 and all additional Python packages installed as described in the Beta 0.2.4 release of STREAMLINE \cite{streamline}.

\section{Results and Discussion}
\label{sec:3}
In this section we summarize high level findings of STREAMLINE experiments on the datasets described in the previous section. As should be apparent from the description of STREAMLINE above, there are many results and figures that could be presented, therefore instead of including them here we have made the STREAMLINE PDF report summaries for each 'experiment' available at: \url{https://github.com/UrbsLab/STREAMLINE/tree/main/Experiments}. Documentation of all pipeline settings for respective experiments are included on the first page of each report.

\subsection{HCC UCI Benchmark Data Results}
\begin{figure}[t!]
    \centering
    \includegraphics[width = \textwidth]{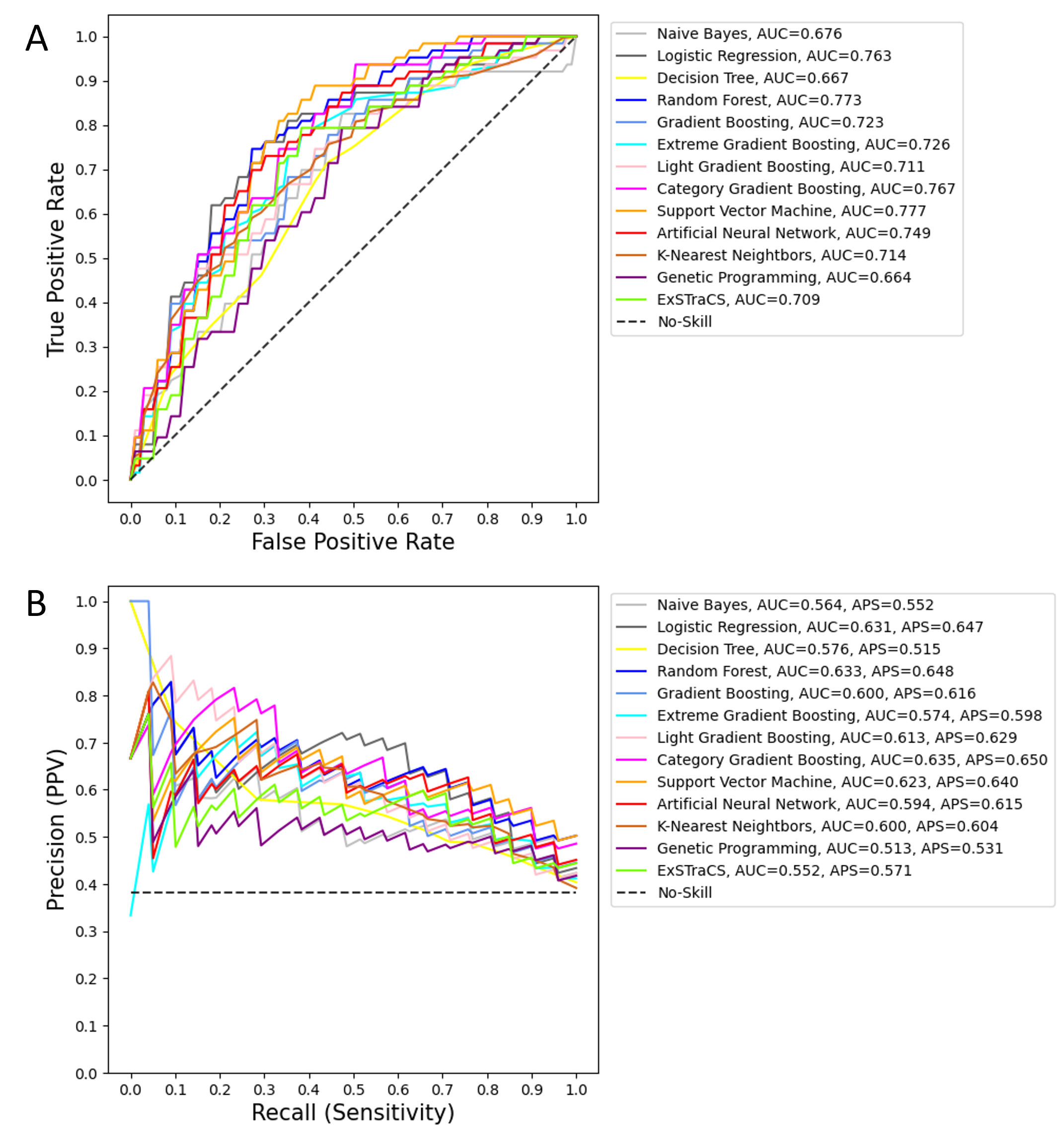}
    \caption{Full HCC dataset analysis: (A) ROC plot, (B) PRC plot, each illustrating averages over all CV models.}
    \label{hcc_curves}
\end{figure}

This analysis was designed to illustrate the overall functionality of the entire pipeline on a small, simple dataset. We repeated this STREAMLINE experiment on the same set of two HCC datasets with the same pipeline configurations and random seed to confirm complete reproducability (see aforementioned PDF experiment summaries on GitHub). Focusing on the original HCC dataset, Figure \ref{hcc_curves} compares ROC and PRC ML algorithm performance. While SVM yielded the highest average ROC-AUC and CGB yielded the highest average PRC-AUC Kruskal-Wallis testing indicated these differences were not significant, likely in part due to the small sample size available in this analysis. A review of mean and median metric performance across all algorithms highlights how there is no single clear algorithmic 'winner' across all metrics. 

%Figure 10

\begin{figure}[t!]
    \centering
    \includegraphics[width = \textwidth]{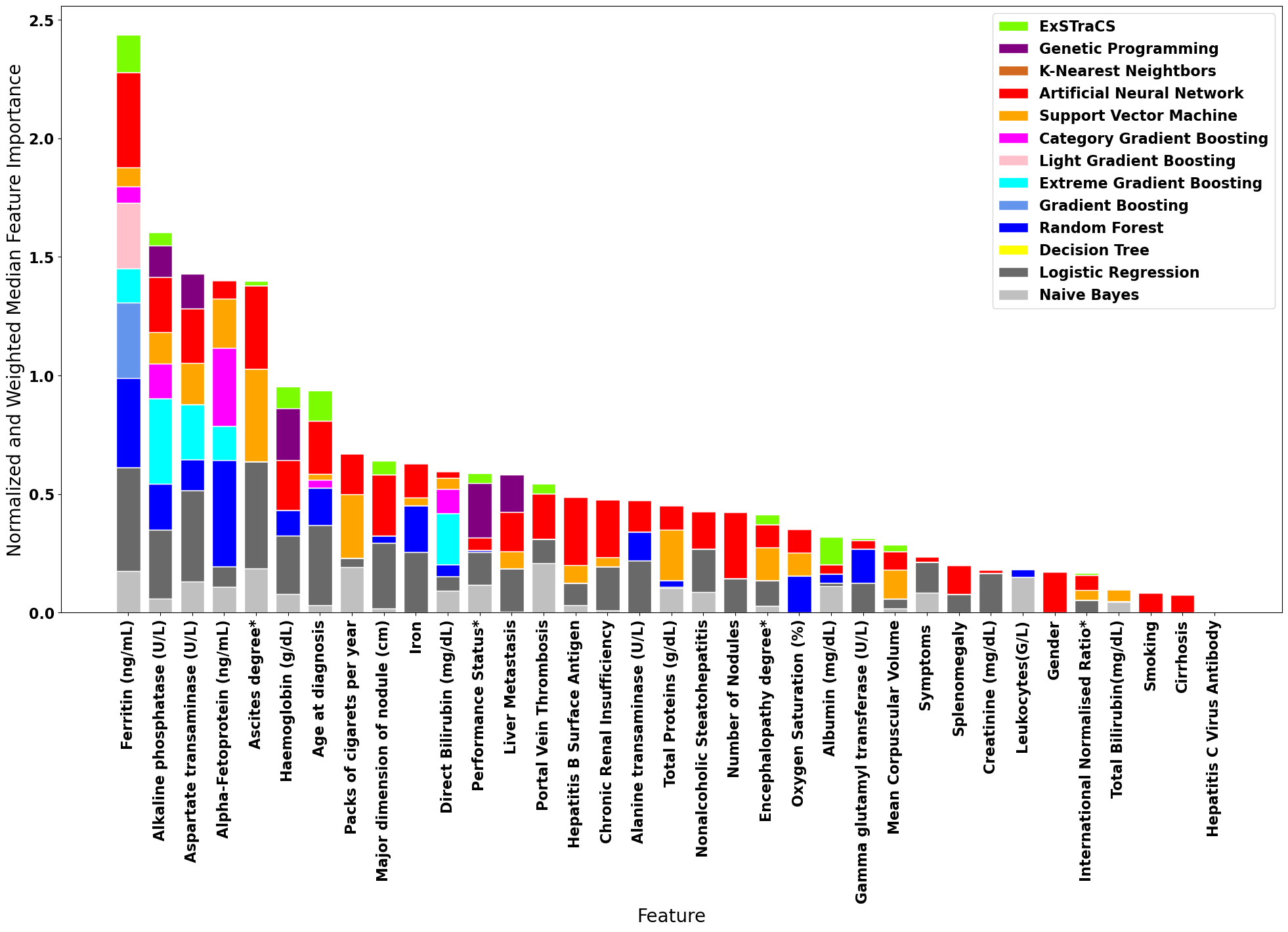}
    \caption{CFIBP of the full HCC dataset (top 40 features).}
    \label{cfibp_hcc}
\end{figure}

Figure \ref{cfibp_hcc} gives model feature importance estimates across all ML algorithms, indicating the feature, Ferritin (ng/ML), to be most consistently informative. In contrast with pre-modeling feature importance estimates, this feature had previously only been ranked 6th by MI 11th by MultiSURF.

%Figure 11

\begin{figure}[h]
    \centering
    \includegraphics[width = 0.75\textwidth]{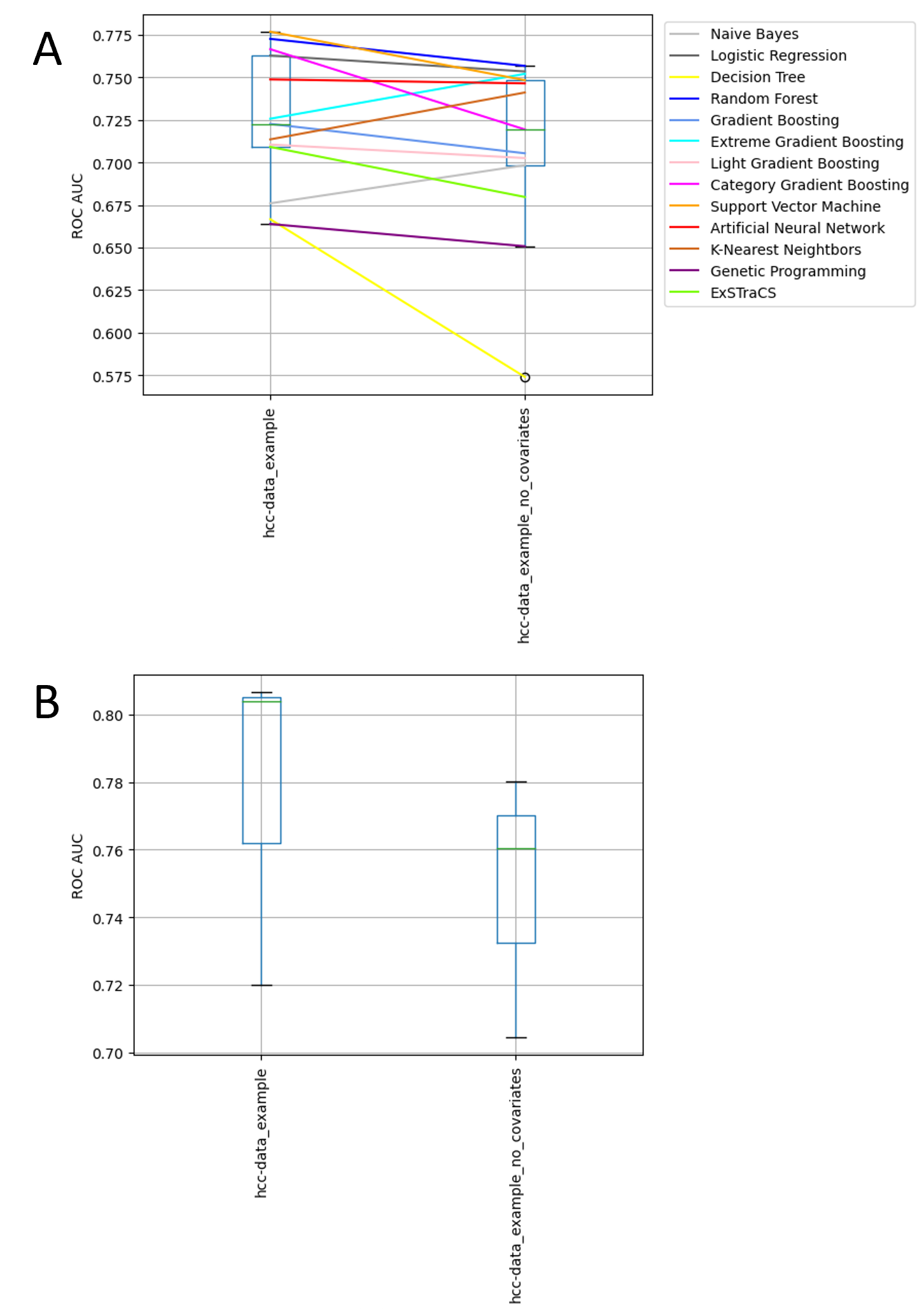}
    \caption{Comparing performance on HCC data with and without covariates. (A) Comparing average ML ROC-AUC between all algorithms. The average performance shift of each algorithm is indicated by the respective line connecting the boxplots. (B) Compares individual CV runs of SVM which yielded the best ROC-AUC on the full HCC dataset. A clear but non-significant reduction in performance was observed when covariates were removed.}
    \label{hcc_compare}
\end{figure}

Lastly, Figure \ref{hcc_compare} compares performance of ML algorithm between the full HCC dataset and the same data with the covariate features removed. This pairing of boxplots illustrates how investigators can use STREAMLINE output to look more closely at specific performance differences across and within ML algorithms and datasets.

%Figure 12

\subsection{GAMETES Simulated Datasets Results}
\begin{figure}[h]
    \centering
    \includegraphics[width = \textwidth]{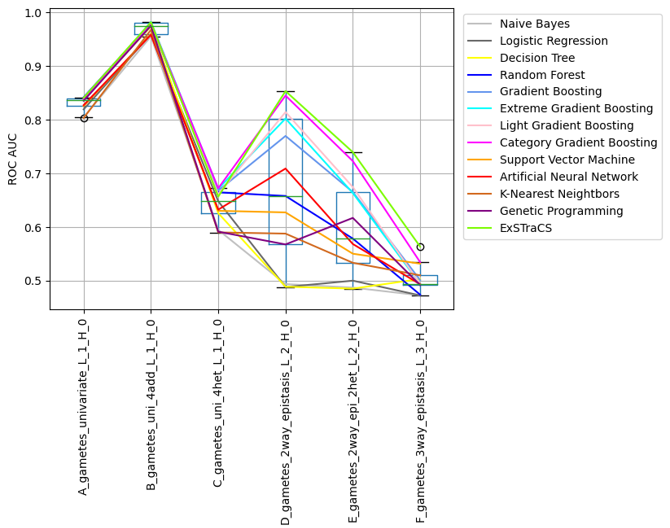}
    \caption{Comparing average ML ROC-AUC performance across GAMETES datasets (A-F).}
    \label{gametes_compare}
\end{figure}

This analysis was designed to highlight basic advantages and disadvantages of ML algorithms in different GAMETES-simulated data scenarios. Figure \ref{gametes_compare} compares average ROC-AUC performance across all 6 GAMETES simulations. We observe fairly consistent performance across algorithms for the first two 'easiest' datasets, and start to observe increasingly dramatic performance differences as heterogeneity and feature interactions are introduced. 

%Figure 13

We focus on mean ROC-AUC in the following performance discussion. For dataset (A), i.e. univariate, all algorithms performed similarly well, with KNN performing least well. For dataset (B), i.e. additive univariate, all algorithms performed similarly well, with NB performing least well. For dataset (C), i.e. heterogeneous univariate, CGB, XGB, RF, GB, and ExSTraCS performed best, while NB, KNN, and GP performed least well. For dataset (D), i.e. pure 2-way epistasis, we observe the first dramatic performance differences with ExSTraCS, CGB, LGB, XGB, and GB performing similarly at the top, while NB, LR, and DT entirely failed to detect the 2-feature interaction. For dataset (E), i.e. heterogeneous pure 2-way epistasis, ExSTraCS and CGB stood out as top performers, while again NB, LR, and DT failed. Lastly, for dataset (F), i.e. pure 3-way epistasis, all algorithms struggled with this noisy complex association, but ExSTraCS performed slightly but significantly better across a number of metrics. Across these simulated scenarios, ExSTraCS performed consistently well and often best of all ML algorithms included highlighting the competitive efficacy of evolutionary rule-based machine learning. However, as would be expected, no algorithm stands out as being ideal across all data scenarios, or evaluation metrics.

For all datasets, model feature importance CFIBP's correctly differentiated simulated relevant from irrelevant features.   

\subsection{\emph{x}-bit MUX Benchmark Datasets Results}
\begin{figure}[h]
    \centering
    \includegraphics[width = \textwidth]{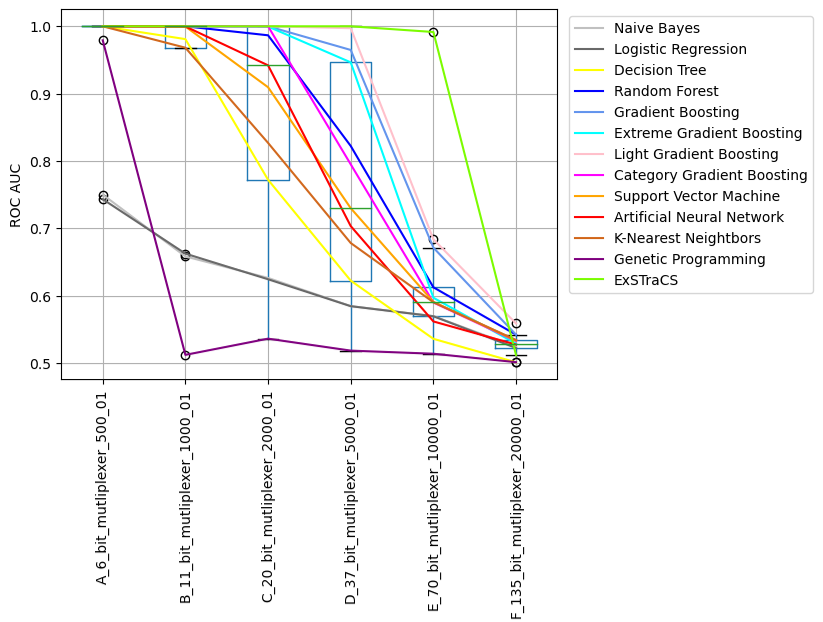}
    \caption{Comparing average ML ROC-AUC performance across 6, 11, 20, 37, 70,and 135-bit multiplexer benchmark datasets.}
    \label{multi_compare}
\end{figure}

\begin{figure}[h]
    \centering
    \includegraphics[width = \textwidth]{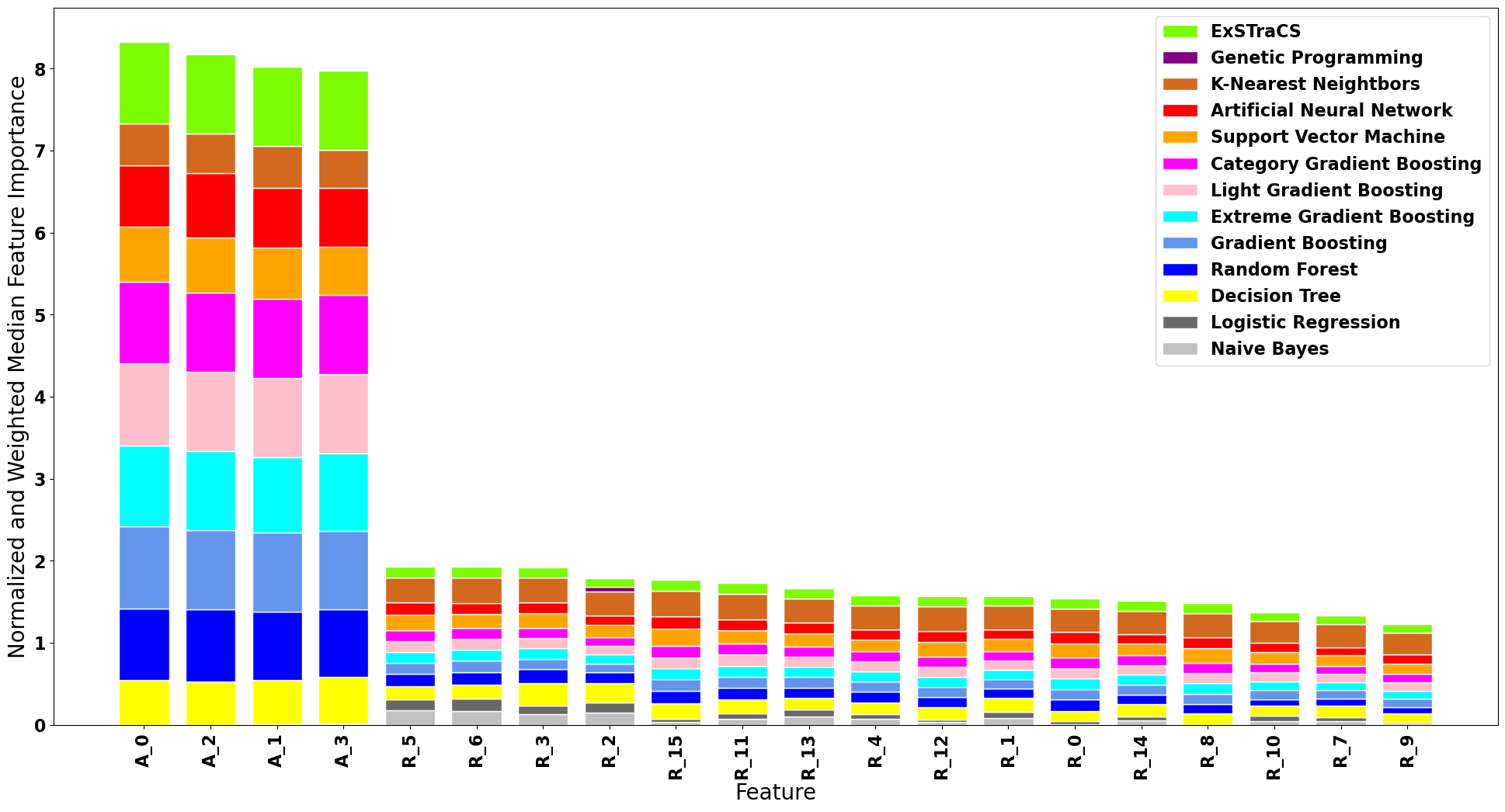}
    \caption{CFIBP of 20-bit MUX. This dataset includes 4 address bits `A' and 16 register bits `R'. In MUX problems, address bits are relevant in predicting all instances, but register bits are only relevant in the subset of instances where the address bits point to them. This is why address bits have a much larger feature importance than register bits in this plot. In the 20-bit MUX, 16 separate 5-way interactions, each involving 4 address and 1 register bit must be found to solve the problem.}
    \label{multi_cfibp}
\end{figure}

This analysis was designed to examine how different ML algorithms performed when modeling data with an increasing underlying problem complexity as well as to highlight how other well-known ML algorithms compare to ExSTraCS (previously shown to perform well on these benchmark problems). Figure \ref{multi_compare} compares performance across all 6 MUX datasets. The simplest, 6-bit MUX is solved perfectly (i.e. ROC-AUC = 1, averaged over CV partitions) by all algorithms except NB, LR, and GP, although GP achieved an ROC-AUC of 0.979. These three algorithms failed to solve all subsequent MUX datasets. For the 11-bit MUX, only the ensemble tree-based ML algorithms (i.e. RF, GB, XGB, LGB, and CGB) and ExSTraCS performed perfectly, and other algorithms yielded deteriorating performance. GP in particular yielded a dramatic performance drop between the 6 and 11-bit MUX. For the 20-bit MUX, only CGB and ExSTraCS yield perfect performance, although GB, XGB, LGB yielded near perfect performance, and RF achieves an ROC-AUC of 0.987. Figure \ref{multi_cfibp} presents the CFIBP for the 20-bit MUX, highlighting the expected increased feature importance estimates for address bit in contrast with register bit features. For the 37-bit MUX, only ExSTraCS yielded perfect performance, although GB, XGB, and LGB yielded ROC-AUC's over 0.9.  For the 70-bit MUX, ExSTraCS performed dramatically better than all other algorithms with an ROC-AUC of 0.992. Lastly for the 135-bit MUX, all algorithms failed to perform well, with LGB performing best with an ROC-AUC of 0.559. This failure was largely expected as only 18K instances were available for model training, when previous work required 40K instances for ExSTraCS to closely solve the 135-bit MUX where a staggering $4.36e40$ unique binary instances make up the problem space. In this particular benchmark analysis, ExSTraCS stands out clearly as the best performing algorithm. Overall this analysis highlights the capability of RBML algorithms such as ExSTraCS to perform competitively, and sometimes better than other well-established ML modeling approaches. Additionally, it provides another example of how STREAMLINE facilitates comparison of ML algorithm performance as a structured, rigorous analysis.

%Figure 14

%Figure 15

\section{Conclusions}
\label{sec:4}
In this paper we introduced STREAMLINE, a simple transparent end-to-end AutoML pipeline. Initial application of STREAMLINE to a variety of benchmark datasets illustrates how this tool can be used beyond a standard data analysis to facilitate comparison of (1) new to established ML modeling algorithms, (2) ML performance across different target datasets, (3) other autoML tools to rigorous benchmark. In this first release, we sought to include a mix of well known ML algorithms and a handful of newer or less well known algorithms with unique advantages (i.e. genetic programming and rule-based machine learning). These findings in applying STREAMLINE support the importance of utilizing a variety ML modeling algorithms, each with different strengths and weaknesses. 

Future work will target many expansions and new applications for STREAMLINE including (1) extending the framework to accommodate multi-class and regression analysis, (2), adding new algorithms, visualizations, and automated elements to the pipeline including one-hot encoding for categorical variables and Shapley values for model interpretation, (3) providing support to run STREAMLINE in different parallel computing environments, (4) adding environment controls such as 'Docker' to further facilitate reproducibility, (5) applying it to comprehensively evaluate new feature selection and ML modeling methodologies, (6) applying it as a positive control benchmark to evaluate potential performance advantages using other AutoML or artificial-intelligence-driven ML tools, (7) identifying a subset of ML algorithms that can be reliably applied in different scenarios to enhance understanding of a given dataset, (8) leveraging this platform as an educational and algorithm development tool by facilitating overall analysis comprehension and transparency and (8) utilize the STREAMLINE framework to encourage a broader adoption of promising, less-well known strategies such as GP or rule-based ML in the future.

\section*{Acknowledgements}
The study was supported by the following NIH grants: R01s LM010098 and AG066833.  STREAMLINE development benefited from multiple biomedical research collaborators at the University of Pennsylvania, Fox Chase Cancer Center, Cedars Sinai Medical Center, and the University of Kansas Medical Center. Special thanks to Patryk Orzechowski, Trang Le, Sy Hwang, Richard Zhang, Wilson Zhang, and Pedro Ribeiro for their code contributions and feedback. We also thank the following collaborators for their feedback on application of the pipeline during development: Shannon Lynch, Rachael Stolzenberg-Solomon, Ulysses Magalang, Allan Pack, Brendan Keenan, Danielle Mowery, Jason Moore, and Diego Mazzotti.

\bibliographystyle{unsrt}  
\bibliography{streamline}  %%% Remove comment to use the external .bib file (using bibtex).

%%%\begin{thebibliography}{1}

%%%\bibitem{hadash2018estimate}
%%%Guy Hadash, Einat Kermany, Boaz Carmeli, Ofer Lavi, George Kour, and Alon
%%%  Jacovi.
%%%\newblock Estimate and replace: A novel approach to integrating deep neural
%%%  networks with existing applications.
%%%\newblock {\em arXiv preprint arXiv:1804.09028}, 2018.

%%%\end{thebibliography}

\end{document}